\documentclass{isprs} 
\usepackage{setspace}
\usepackage{geometry} 
\usepackage{epstopdf}
\usepackage[labelsep=period]{caption}  
\usepackage[british]{babel} 
\usepackage[hang]{footmisc}

\usepackage{microtype}
\usepackage{graphicx}
\usepackage{booktabs}
\usepackage{xurl}

\usepackage{dblfloatfix}

\usepackage{times}
\usepackage{epsfig}
\usepackage{graphicx}
\usepackage{amsmath}
\usepackage{amssymb}
\usepackage[utf8]{inputenc}
\usepackage{subcaption}
\usepackage{tumabbrev}

\usepackage{booktabs}
\usepackage{multirow}

\usepackage{appendix}

\usepackage{cleveref}

\usepackage{tikz}
\usepackage{pgfplots}
\usetikzlibrary{positioning, calc,arrows,arrows.meta, fit}
\usepgfplotslibrary{groupplots}
\usepgfplotslibrary{fillbetween}
\usepgfplotslibrary{statistics} 
\usepackage{xfrac}

\usetikzlibrary{shapes,snakes}

\usepackage{tumcolors}
\usepackage{tummath}

\newcommand{\f}{f}

\usepackage[super]{nth}
\usepackage{mathtools}

\definecolor{evalcolor}{HTML}{3F3F3F}
\definecolor{traincolor}{HTML}{B98951}
\definecolor{validcolor}{HTML}{3F4BBE}

\colorlet{colortrain}{tumblue}
\colorlet{colorinfer}{tumblack}

\colorlet{earlinesscolor}{tumblue}
\colorlet{accuracycolor}{tumorange}

\colorlet{stdcolor}{tumbluelight}
\colorlet{mediancolor}{tumorange}
\colorlet{meancolor}{tumblue}

\colorlet{frh01color}{tumgray}
\colorlet{frh02color}{tumorange}
\colorlet{frh03color}{tumblue}
\colorlet{frh04color}{tumblack}

\colorlet{b1color}{tumdiagramaubergine}
\colorlet{b2color}{tumdiagramnavyblue}
\colorlet{b3color}{tumdiagramturquoise}
\colorlet{b4color}{tumdiagramgreen}
\colorlet{b5color}{tumdiagramlimegreen}
\colorlet{b6color}{tumdiagramyellow}
\colorlet{b7color}{tumdiagramsand}
\colorlet{b8color}{tumdiagramredorange}
\colorlet{b8Acolor}{tumdiagramred}
\colorlet{b9color}{tumblack}
\colorlet{b10color}{tumblue}
\colorlet{b11color}{tumdiagramdarkred}
\colorlet{b12color}{tumorange}

\colorlet{epsilon0color}{tumorange}
\colorlet{epsilon1color}{tumblue}
\colorlet{epsilon10color}{tumblack}

\colorlet{meadowcolor}{tumbluemedium}
\colorlet{wbarleycolor}{tumbluedark}
\colorlet{corncolor}{tumorange}
\colorlet{wheatcolor}{tumgreen}
\colorlet{sbarleycolor}{tumdiagramred}
\colorlet{clovercolor}{tumdiagramturquoise}
\colorlet{triticalecolor}{tumdiagramsand}

\tikzstyle{rnn}=[draw,circle, inner sep=.1em]
\tikzstyle{norm}=[rounded corners,draw]
\tikzstyle{annot}=[rounded corners, fill=tumblue!20]
\tikzstyle{infer}=[-stealth, shorten >=.0em, shorten <=.0em, colorinfer]
\tikzstyle{loss}=[fill=tumblue!10, rounded corners, font=\small]
\tikzstyle{grad}=[colortrain]

\tikzstyle{test} = [thick]
\tikzstyle{train} = [thin, dotted]

\usepackage[inline]{enumitem}
\setenumerate{label=(\roman*),itemsep=3pt,topsep=3pt}

\usetikzlibrary{external,pgfplots.dateplot}
\tikzexternaldisable

\usepackage[eulergreek]{sansmath}
\pgfplotsset{
	y tick label style={/pgf/number format/.cd,%
		scaled y ticks = false,
		set thousands separator={},
		fixed},
	x tick label style={/pgf/number format/.cd,%
		scaled x ticks = false,
		set decimal separator={,},
		fixed},
	tick label style = {font=\scriptsize\sansmath\sffamily},
	every axis label = {
		font=\scriptsize\sansmath\sffamily},
	every axis/.append style={
		axis lines=left, 
		enlargelimits, 
		thick},
	legend style = {font=\scriptsize\sansmath\sffamily, draw=none, rounded corners, fill opacity=.5, text opacity=1},
	label style = {font=\scriptsize\sansmath\sffamily},
	grid style={line width=.1pt, draw=gray!10},
	major grid style={line width=.2pt,draw=tumgraylight},
}

\tikzstyle{circ} = [circle, draw=white, fill=tumblue, inner sep=1pt]

\tikzstyle{druschdatum} = [thin, star,star points=3, star point ratio=0.5, inner sep=.15em, draw=tumwhite, fill=tumblue]

\usepackage[frozencache,cachedir=.]{minted} 
\usepackage{siunitx}

\usepackage{enumitem,amssymb}
\newlist{todolist}{itemize}{2}
\setlist[todolist]{label=$\square$}
\usepackage{pifont}

\newcommand{\ourdataset}{\emph{BreizhCrops}\xspace}
\newcommand{\gitrepo}{\url{https://github.com/dl4sits/breizhcrops}}

\begin{document}

\title{BreizhCrops: A Time Series Dataset for Crop Type Mapping}

\author{Marc Ru\ss{}wurm\textsuperscript{1 }, Charlotte Pelletier\textsuperscript{2 }, Maximilian Zollner\textsuperscript{1 }, Sébastien Lefèvre\textsuperscript{2 }, Marco Körner\textsuperscript{1 }}

\address{
	\textsuperscript{1 }Chair of Remote Sensing Technology, Department of Aerospace and Geodesy, Technical University of Munich, Germany\\
	\textsuperscript{2 }
	Univ. Bretagne Sud, UMR 6074, IRISA, F-56000 Vannes, France\\
}


\commission{}{} 
\workinggroup{} 
\icwg{}   


\abstract{
We present \ourdataset, a novel benchmark dataset for the supervised classification of field crops from satellite time series. We aggregated label data and Sentinel-2 top-of-atmosphere as well as bottom-of-atmosphere time series in the region of Brittany (Breizh in local language), north-east France. 
We compare seven recently proposed deep neural networks along with a Random Forest baseline. 
The dataset, model (re-)implementations and pre-trained model weights are available at the associated GitHub repository (\gitrepo) that has been designed with applicability for practitioners in mind.
We plan to maintain the repository with additional data and welcome contributions of novel methods to build a state-of-the-art benchmark on methods for crop type mapping.
}

\keywords{Satellite Image Time Series, Deep Learning, Crop Type Mapping, Dataset, Benchmark, Sentinel-2}

\maketitle

\section{Introduction}

The Earth's surface is governed by spatio-temporal processes that are measured by various satellites on discrete temporal intervals.
Extracting knowledge from this data at a large scale is a key objective to modern remote sensing research. 
The related field of machine learning has demonstrated that direct model comparisons through application-specific benchmarks are a central driver for rapid development in the field.
So far, a direct comparison of models has been difficult in remote sensing due to the diverse nature of remote sensing data, the partly proprietary access to satellite data and labels, and the exclusive expertise on data-processing in remote sensing.
Hence, proposed time series methods have been tested predominately on self-compiled datasets rather than on a common benchmark.
The spatial component in remote sensing data has been explored through its methodological relation to computer vision and a variety of related benchmark datasets have been proposed, such as \texttt{DeepGlobe2018}~\cite{DeepGlobe18}, \texttt{SEN12MS}~\cite{schmitt2019sen12ms} or \texttt{BigEarthNet}~\cite{sumbul2019bigearthnet}.
The temporal component has received less attention. 
MediaEval Benchmarking Initiative has recently run a task on Emergency Response for Flooding Events that included imagery from multiple dates\footnote{http://www.multimediaeval.org/mediaeval2019/multimediasatellite/} and a benchmark dataset for change detection has been proposed~\cite{daudt2018urban}.
For the temporal task of crop type mapping from satellite image time series, novel approaches have predominantly been tested on self-created datasets and only partly compared to other state-of-the-art methods \cite{pelletier2019temporal,russwurm2017multi,russwurm2018multi}.
Time series datasets involving land cover classification labels have been proposed~\cite{tiselac,gee-tsda}. However, and to the best of our knowledge, no public benchmark for satellite time series classification that comprehensively compares existing models is available to this date.

In this work, we propose a novel large-scale satellite image time series dataset for crop type mapping termed \ourdataset from the region of Brittany, France. 
We extracted time series from Sentinel-2 at two processing levels (top- and bottom-of-atmosphere) which yielded more than 600k multivariate time series examples each. Each time series sample is labeled with one of nine crop type classes. We then use \ourdataset to benchmark a series of seven classification algorithms including Random Forest and six deep learning methods based either on convolution, recurrence, or attention.
A first version of this dataset has been presented at the \emph{Time Series Workshop} at the \emph{International Conference on Machine Learning (ICML) 2019} in a contributed talk. 

\begin{figure}[t]
  
  \begin{subfigure}[t]{.49\linewidth}
  \includegraphics[width=\textwidth]{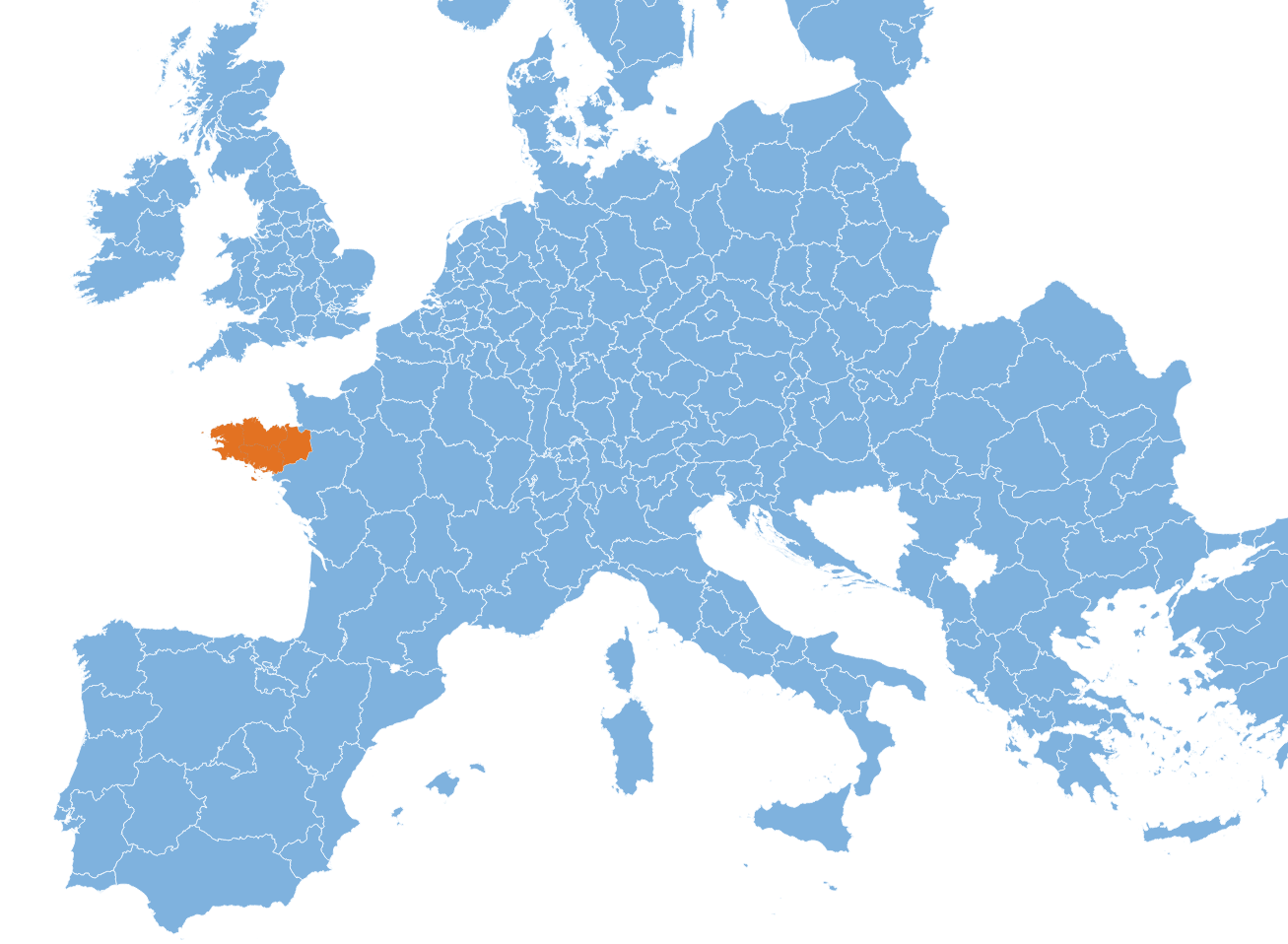}
  \caption{Brittany within the NUTS-2 regions in Europe}
  \label{fig:aoi:europe}
  \end{subfigure} 
  \begin{subfigure}[t]{.49\linewidth}
    \includegraphics[width=\textwidth]{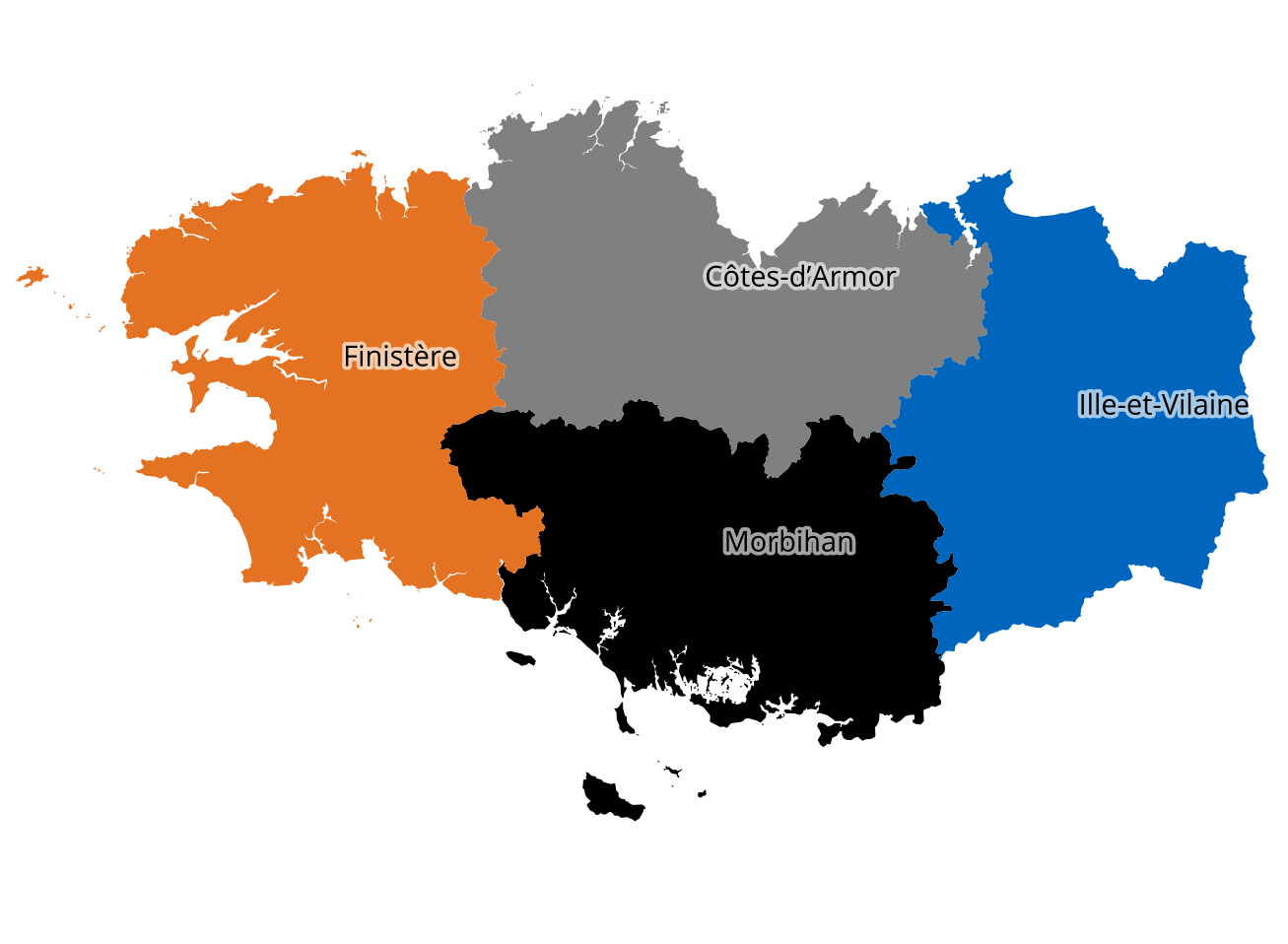}
    \caption{The NUTS-3 departments of Brittany used for data partitioning.}
    \label{fig:aoi:brittany}
  \end{subfigure}

  \caption{The NUTS-3 region \emph{FRH0} of Brittany, France.}
  \label{fig:aoi}
 
\end{figure}

The remainder of this paper is organized as follows: \cref{sec:dataset} presents the \ourdataset dataset. Then \cref{sec:models} briefly describes the seven classification algorithms that are evaluated and compared in \cref{sec:expe}. We then discuss some challenges that can be addressed by using \ourdataset in \cref{sec:challenges}. \Cref{sec:code} provides a minimal working example to download \ourdataset. Finally, we draw the conclusions in \cref{sec:conlusion}.

\section{The BreizhCrops Dataset}
\label{sec:dataset}

The studied area is the Brittany region (Breizh in the local language) located in the northwest of France and covering 27,200 km², as shown in \cref{fig:aoi}. The region is dominated by a temperate oceanic climate (K\"{o}ppen classification) with an annual average temperature ranging from 5.6\textdegree\ in winter to 17.5\textdegree\ in summer and mean annual precipitation of 650 millimeters.

The dataset comprises about 610k labeled observations per processing level. 
Each observation describes the temporal profile of a field crop and corresponds to a multivariate time series obtained by averaging at the crop field level reflectance values extracted from Sentinel-2 images. 
%
%



\begin{figure*}
  \begin{subtable}[b]{.5\textwidth}
    \small
    \begin{tabular}{@{}lrrrr@{}}
      \toprule
      Departments & NUTS-3 & Parcels & \# L1C & \# L2A \\
 
      \cmidrule(r){1-1}\cmidrule(lr){2-2}\cmidrule(l){3-3}\cmidrule(l){4-4}\cmidrule(l){5-5}
      Côtes-d’Armor & FRH01 & 221,095 & 178,613 & 178,632\\
      Finistère & FRH02 & 180,565 & 140,645 & 140,782\\
      Ille-et-Vilaine & FRH03 & 207,993 & 166,391 & 166,367\\
      Morbihan & FRH04 & 158,522 & 122,614 & 122,708\\
      \midrule
      Total & & 768,175 & 608,263 & 608,489\\
      \bottomrule
    \end{tabular}

    \caption{NUTS-3 departments of Brittany with number of field parcels and time series for each processing level.}
    \label{fig:class:region}
  \end{subtable}
  \begin{subfigure}[b]{.5\textwidth}
    \tikzsetnextfilename{partition_histograms}
\begin{tikzpicture}
  \begin{axis}[
        ybar, 
        axis on top,
        title={},
        height=3.5cm, width=1\linewidth,
        bar width=1mm,
        ymajorgrids, 
        tick align=outside,
        major grid style={draw=tumwhite},
        ymin=0,
        ymode=log,
        axis line style={opacity=1, thin},
        enlarge x limits=.05,
        legend style={
            at={(1,1.5)},
            anchor=north east,
            draw=none,
            legend columns=2,
            rounded corners=0,
            /tikz/every even column/.append style={column sep=0.5cm, font=\scriptsize}
        },
        ylabel={Number of Parcels},
        ylabel style={yshift=-1.3em},
        xtick={0,1,...,12},
        xticklabels={barley, wheat, rapeseed, corn, sunflower, orchards, nuts, perm. meadows, temp. meadows},
        tick label style={rotate=20,anchor=east}
    ]
    \addplot [draw=none, fill=frh01color] table[x=id,y=frh01, col sep=comma] {images/l1c_counts.csv};
    \addplot [draw=none, fill=frh02color] table[x=id,y=frh02, col sep=comma] {images/l1c_counts.csv};
    \addplot [draw=none, fill=frh03color] table[x=id,y=frh03, col sep=comma] {images/l1c_counts.csv};
    \addplot [draw=none, fill=frh04color] table[x=id,y=frh04, col sep=comma] {images/l1c_counts.csv};

    \legend{Côtes-d’Armor {\tiny(FRH01)},Finistère {\tiny(FRH02)},Ille-et-Vilaine {\tiny(FRH03)},Morbihan {\tiny(FRH04)}}
  \end{axis}
  \end{tikzpicture}
    \caption{The class frequencies of the nine selected crop types show an imbalance of common crops which is a frequent problem of crop type mapping.}
    \label{fig:classfrequencies}
  \end{subfigure}
  \caption{Analyses of the number of parcels and class frequencies per partition in the vector dataset.}
\end{figure*}
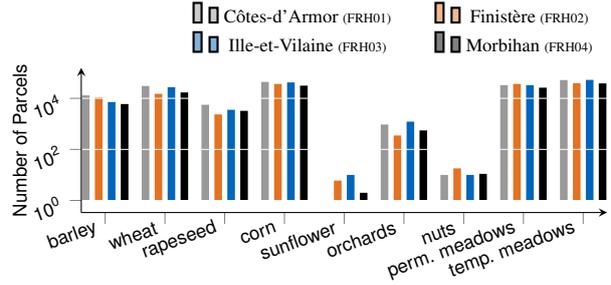

\subsection{Crop Type Labels}

The \emph{Common Agricultural Policy} of the European Union subsidizes farmers based on their cultivated crops.
Each member country is required to gather geographical information about the geometry and cultivated crops. 
This information is obtained from the farmers themselves by surveys within the subsidy application process.
National agencies monitor the correctness either by gathering control samples \textit{in-situ} or by means of remote sensing and Earth observation.
In France, the \emph{National Institute of Forest and Geography Information (IGN)} is responsible for gathering this information, the so-called Agricultural Land Parcel Information System (\textit{Registre Parcellaire Graphique})---RPG.
The IGN institute recently started releasing anonymized parcel geometries and types of cultivated crops with an \emph{open license} policy\footnote{\url{https://www.data.gouv.fr/en/datasets/registre-parcellaire-graphique-rpg-contours-des-parcelles-et-ilots-culturaux-et-leur-groupe-de-cultures-majoritaire}}. 

The raw crop type categories contain 328 unique crop labels grouped into 23 groups. For \ourdataset dataset, we selected the 9 following crop categories: \emph{barley}, \emph{wheat}, \emph{rapeseed}, \emph{corn}, \emph{sunflower}, \emph{orchards}, \emph{nuts}, \emph{permanent meadows} and \emph{temporary meadows}. The field labels have been gathered in the year 2017.
We decided to keep ``well-defined'' classes and avoided broad categories, such as \emph{diverse} or \emph{fodder crops}.
We also made the choice of keeping two minority classes (\emph{sunflower} and \emph{nuts}), which have only a few occurrences in RPG, as a challenge to the classification models. It also reflects the strong class imbalance in real-world crop-type-mapping datasets, as shown in \cref{fig:classfrequencies}.

\subsection{Satellite Data}

The dataset is composed of Sentinel-2 image time series extracted from January 1, 2017 to December 31, 2017\footnote{Note that Sentinel-2B images are available around July 2017 as this second Sentinel-2 satellite was launched in March 2017.}.
We aggregated the satellite time series data in two processing levels: the raw reflectances at the top-of-atmosphere (level  1C) and the atmospherically corrected surface reflectances at the bottom-of-atmosphere (level 2A). 

For both processing levels, we average reflectance values over the bounds of the field geometry retrieved from the dataset. Each spectral band is mean-aggregated over one field parcel to a feature vector $\V{x}_t \in \R^{D}$, with $D$ the number of features and $t$ a timestamp.
This aggregation strategy requires known field geometries that are accessible for most of the fields in Europe~\cite{leo2001land}.
In case no geometry data is available, a trained model can be inferred with feature vectors from each pixel of a Sentinel-2 image time series. 


\cref{fig:example} displays examples of the multivariate time series given as inputs to the classification models. \Cref{fig:example:meadows,fig:example:corn} show the satellite time series of a corn and meadow example at processing level L1C with 13 spectral bands, as provided to the classifiers.
The data is positively biased in single observations by clouds which cause systematically positive outliers values in the time series data.
\Cref{fig:examplel2a:corn} shows the same corn parcel example at the L2A processing level with 10 remaining spectral bands.

\subsubsection{Top-of-Atmosphere}

We chose to include L1C top-of-atmosphere due to the ease of adoption of methods to other regions where the access to atmospherically corrected data is not guaranteed.

To obtain the top-of-atmosphere satellite data, we downloaded \emph{all} available \emph{Sentinel~2} images (without filtering on the cloud coverage) at processing level L1C from \emph{Google Earth Engine} (GEE) \cite{gorelick2017google}. This resulted in either 51 or 102 observations per field parcel, as shown in \cref{fig:example:meadows,fig:example:corn}. 

\subsubsection{Bottom-of-Atmosphere}

\begin{figure}
  
  \input{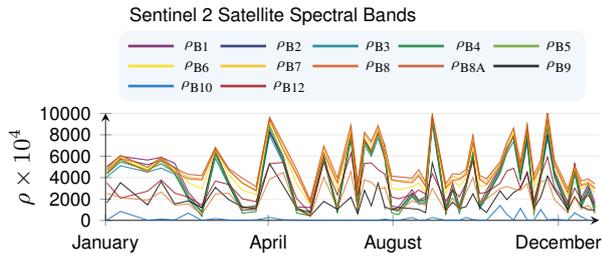}
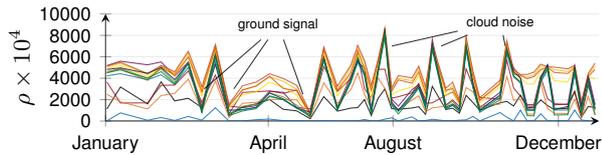
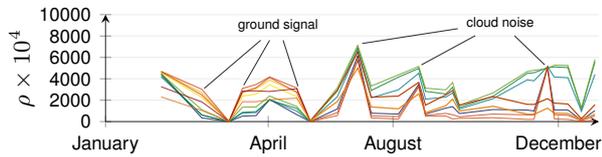
  \begin{subfigure}{.5\textwidth}
    \examplemeadows
    \caption{A single example of category \textsl{temporary meadows} (L1C)}
    \label{fig:example:meadows}
  \end{subfigure}
  
  \begin{subfigure}{.5\textwidth}
    \examplecorn
    \caption{A single example of category \textsl{corn} (L1C)}
    \label{fig:example:corn}
  \end{subfigure}
  
  \begin{subfigure}{.5\textwidth}
    \examplecornla
    \caption{A single example of category \textsl{corn} (L2A)}
    \label{fig:examplel2a:corn}
  \end{subfigure}
  
  \caption{Examples of the input time series of top-of-atmosphere reflectances $\rho$ for all 13 spectral bands of the Sentinel-2 satellite.}
  \label{fig:example}

\end{figure}

We include L2A bottom-of-atmosphere data where images acquired over time and space share the same reflectance scale. Satellite images corrected from atmospheric effect might improve land cover mapping when monitoring large scale areas over time~\cite{song2001classification}.
In total, we downloaded 374 images (over the seven Sentinel-2 tiles that covered Brittany) that are corrected from atmospheric, adjacency and slope effects by MAJA processing chain~\cite{hagolle_2015} from PEPS --~a French portal for Sentinel-2 data\footnote{\url{https://peps.cnes.fr}}. Only images with a cloud-cover below 80~\% are processed by MAJA. Hence, we collected an average of 53 images per tile.
We show an example of a field parcel in \cref{fig:examplel2a:corn}.

After MAJA processing, only 10 spectral bands are available--~the three Sentinel-2 spectral bands at a 60-meter spatial resolution serve only to apply the atmospheric correction and to detect clouds. 

Note that the process of several Sentinel-2 tiles results in samples representing reflectance values on different timestamps that might be disturbed by clouds (and shadows). This temporal sampling inhomogeneity between observation requires usually an additional preprocessing such as an interpolation, a subsampling or padding.

\subsection{Data organisation}
The data is organized at a regional level by the \emph{Nomenclature of Territorial Units for Statistics (NUTS)} which forms the European standard for referencing authoritative districts.
Brittany is the NUTS-2 region FRH0, as highlighted in \cref{fig:aoi:europe}.
It is further divided into the four NUTS-3 regions (\cref{fig:aoi:brittany}): \emph{Côtes-d’Armor} (FRH01), \emph{Finistère} (FRH02), \emph{Ille-et-Vilaine} (FRH03), and \emph{Morbihan} (FRH04).
We partitioned all acquired field parcels according to the NUTS-3 regions and suggest to subdivide the dataset into training (FRH01, FRH02), validation (FRH03), and evaluation (FRH04) subsets based on these spatially distinct regions.

\section{Models}
\label{sec:models}
This section describes briefly the deep learning models used for the benchmark as well as the traditional Random Forest algorithm.

\subsection{Random Forest}

Random Forests are one of the most used shallow algorithms for the classification of satellite image time series \cite{gomez2016optical,belgiu_2016} at large scale \cite{inglada2017operational,defourny2019near}. They are able to handle the high dimensionality of satellite image time series datasets, are robust to some class label noise, and are generally insensitive to the choice of hyperparameters \cite{pelletier_2016,pelletier_2017}.

Random Forests are an ensemble approach that trains a set of binary decision trees \cite{breiman_2001}. Each tree is built by using a bootstrap sample (sampling with replacement the training instances). 
The optimal split at each node is determined using an effectiveness test (usually the maximization of the decrease in node impurity) on only a subset of randomly selected variables. 
Both randomization processes (bootstrap sample and random feature subspace) help to increase the diversity among the decision trees.
The tree construction stops when all the nodes are pure (\textit{i.e.}, all node samples belong to the same class) or when a user-defined criterion is met (\textit{e.g.} a maximal depth or a minimum node size).

\subsection{Convolution-based Deep Learning Models}

A one-dimensional convolutional neural network layer extracts features from a temporal local neighborhood by convolving the input time series with a filter bank learned by gradient descent.
In convolutional neural networks, these convolutions are commonly followed by non-linear activation, pooling, and normalization, forming a cascade of layers where the output of one layer feeds the input of the next. Although 1D-convolutional neural networks have gained interest for general-purpose classification of time series \cite{fawaz2019deep}, they have been used only recently for land cover mapping \cite{zhong2019deep,pelletier2019temporal}.

Among the existing approaches, we compare four different models: Temporal Convolutional Neural Network (TempCNN) \cite{pelletier2019temporal},  Multi Scale 1D Residual Network (MSResNet)\footnote{\url{https://github.com/geekfeiw/Multi-Scale-1D-ResNet}}, InceptionTime \cite{fawaz2019inceptiontime} and Omniscale Convolutional Neural Network (OmniscaleCNN) \cite{tang2020rethinking}.
TempCNN~\cite{pelletier2019temporal}
stacks three convolutional layers with convolution filters of the same size, followed by a dense and softmax layers.
MSResNet applies a first convolutional layer followed by a max-pooling operation. The result is then passed through three branches learning six consecutive convolution filters of different lengths and a final global average pooling. For each branch, residual connections are used every three convolutional layers to limit vanishing and exploding gradient issues. Finally, the results are concatenated and passed through the end of the network composed of fully connected and softmax layers.
InceptionTime~\cite{fawaz2019inceptiontime} is an ensemble of five Inception networks that currently obtained the state-of-the-art deep learning results on 85 classification problems of the UCR archive \cite{dau2019ucr}. Each network is composed of a series of six Inception modules followed by a global average pooling operation and a dense layer with a softmax activation. It also makes the use of residual connections every three Inception modules.
OmniscaleCNN \cite{tang2020rethinking} is composed of three convolutional layers followed by a global average pooling and a dense layer with a softmax activation. Its specificity is to concatenate the outputs of several convolution filters whose length is one plus all the prime numbers between two and a quarter of the time series length.
Please note that only TempCNN architecture has been used for land cover mapping from Sentinel-2 image time series~\cite{pelletier2019temporal}. 

\subsection{Recurrence-based Deep Learning Models}

In Recurrent Neural Networks (RNN), layers process a series of observations sequentially while maintaining a feature representation from the previous context. Gated Recurrent Neural Networks, such as Long Short-Term Memory (LSTM)~\cite{hochreiter1997long}, or Gated Recurrent Units~\cite{chung2014empirical} parameterize this context vector by sub-networks termed \emph{gates} which addressed the problem of vanishing gradient through time.
These recurrent layers can be stacked in multiple cascaded layers where the sequence can be introduced bi-directionally in sequence and reversed-sequence orders.
They have been successfully used in remote sensing applications, especially for land cover mapping \cite{russwurm2017temporal,ienco_2017,ndikumana_2018,sun_2018,minh_2018}.

In our experiments, we compare the LSTM
~\cite{hochreiter1997long} network, as evaluated by \cite{russwurm2017temporal}, and the STAR recurrent neural network 
(StarRNN)~\cite{turkoglu2019gating}. StarRNN is composed of STAckable Recurrent cells that require fewer parameters compared to the LSTM or GRU cells and are designed to avoid vanishing gradient issue when using deep architectures (\ie stacking several STAR cells). Both LSTM and StarRNN architectures have been evaluated on crop type mapping.

\subsection{Attention-based Deep Learning Model}

The first use of the attention principle has been proposed by~\cite{bahdanau2014neural} where an importance-score (attention) was computed to weight each element in a sequence. 
The original formulation of attention was used in conjunction with a recurrent neural network. Self-attention~\cite{Vaswani:transformer} reformulated this importance-weighting in self-contained stackable layers that map an input to a same-sized hidden representation.
Self-attention Transformer models~\cite{Vaswani:transformer}, have been originally developed as sequence-to-sequence encoder-decoder models for language translation. 
For sequence-to-label classification only the encoder network that contains stacked self-attention layers is required.
%
In this work, we use a Transformer model that has been evaluated in~\cite{russwurm2019self} for crop type mapping with top-of-atmosphere satellite time series without cloud filtering. It is able to extract features from specific elements in the time series. 

\section{Experiments}
\label{sec:expe}

This section evaluates and compares the seven models presented in section~\ref{sec:models}. We first detail the experimental settings, then we present the obtained results.

\subsection{Experimental setup}

We first provide setting details for the Random Forest classifier and all deep learning models.

\subsubsection{Random Forest}

Following the lead of \cite{inglada2017operational}, we applied a linear temporal interpolation on a regular temporal grid with a time gap of 5 days. 
The linear temporal interpolation is applied for the non-cloudy values: we use cloud masks available in L1C products for the top-of-atmosphere dataset and we use cloud masks computed by the MAJA processing chain for the bottom-of-atmosphere dataset. After the gap-filling operation, each sample is described by a total of 71 $\times D$ variables where 71 represents the number of interpolated dates, and $D$ the number of spectral features used ($D=13$ for top-of-atmosphere data and $D=10$ for bottom-of-atmosphere data).

For the Random Forests hyperparameters, it has been shown that tuning their values results only in a slight performance improvement \cite{cutler_2007}.
Hence, we used a standard hyperparameter setting \cite{pelletier_2016} without performing a hyperparameter tuning: 500 trees at a maximum depth of 25, a number of randomly selected variables per node equals to the square root of the total number of variables.  To conduct the experiments of Section~\ref{subsec:result}, we used the scikit-learn library (Python).

\subsubsection{Deep learning}

\begin{table}[]
    \centering
    \begin{tabular}{lrrr}
        \toprule
        model & epochs & runtime in [it/s] & \# parameters \\
        \cmidrule(lr){1-1}\cmidrule(lr){2-2}\cmidrule(lr){3-3}\cmidrule(lr){4-4}
        TempCNN         & 11 & 1.25  & 3,199,501 \\
        MSResNet        & 23 & 1.04  & 537,325 \\
        InceptionTime   & 23 & 1.03  & 75,533 \\
        OmniscaleCNN    & 19 & 1.02  & 2,739,737 \\
        LSTM            & 17 & 1.16  & 1,339,431 \\
        StarRNN         & 17 & 1.02  & 72,103 \\
        Transformer     & 26 & 1.20  & 188,429 \\
        \bottomrule
    \end{tabular}
    \caption{Model parameters and runtime in iterations per second on a P100-GPU with a batch size of 1024.}
    \label{tab:runtime}
\end{table}

To obtain fixed-length time series that are required for training deep learning methods with batches, 
we decided to randomly sub-sample each time series to a fixed length of 45 observations for the deep learning models while maintaining the sequential topology.

\begin{table*}[h]
\begin{subtable}{\textwidth}
	\begin{tabular}{rllllllll}
		\toprule
		  & shallow & \multicolumn{4}{c}{convolution} &  \multicolumn{2}{c}{recurrence} & \multicolumn{1}{c}{attention}
		  \\
		  \cmidrule(lr){2-2}\cmidrule(lr){3-6}\cmidrule(lr){7-8}\cmidrule(lr){9-9}
		  
		 FRH04	& {RF} & {TempCNN} & {MS-ResNet} & {InceptionTime} & {OmniscCNN} & {LSTM} & {StarRNN} & {Transformer}  \\
overall accuracy & $0.77$ & $0.79$ & $0.78$ & $0.79$ & $0.77$ & $\textbf{0.80}$ & $0.79$ & $\textbf{0.80}$ \\ 
average accuracy & $0.53$ & $0.55$ & $0.51$ & $0.56$ & $0.53$ & $0.57$ & $0.58$ & $\textbf{0.59}$ \\ 
weighted f-score & $0.75$ & $0.78$ & $0.76$ & $0.78$ & $0.74$ & $\textbf{0.79}$ & $0.77$ & $\textbf{0.79}$ \\ 
kappa-metric & $0.69$ & $0.73$ & $0.70$ & $0.73$ & $0.70$ & $\textbf{0.74}$ & $0.72$ & $\textbf{0.74}$ \\
\midrule
FRH01, 02, 04
\\
overall accuracy & $0.76^{\pm 0.02}$ & $0.79^{\pm 0.02}$ & $0.72^{\pm 0.06}$ & $0.71^{\pm 0.07}$ & $0.79^{\pm 0.01}$ & $0.79^{\pm 0.04}$ & $\textbf{0.80}^{\pm 0.02}$ & $\textbf{0.80}^{\pm 0.01}$ \\ 
average accuracy & $0.52^{\pm 0.01}$ & $0.55^{\pm 0.01}$ & $0.56^{\pm 0.05}$ & $0.52^{\pm 0.04}$ & $0.55^{\pm 0.02}$ & $0.56^{\pm 0.03}$ & $0.57^{\pm 0.01}$ & $\textbf{0.58}^{\pm 0.01}$ \\ 
weighted f-score & $0.75^{\pm 0.03}$ & $0.79^{\pm 0.01}$ & $0.71^{\pm 0.05}$ & $0.70^{\pm 0.08}$ & $0.77^{\pm 0.03}$ & $0.78^{\pm 0.05}$ & $0.78^{\pm 0.02}$ & $\textbf{0.80}^{\pm 0.01}$ \\ 
kappa-metric & $0.69^{\pm 0.03}$ & $0.73^{\pm 0.02}$ & $0.66^{\pm 0.05}$ & $0.63^{\pm 0.09}$ & $0.72^{\pm 0.02}$ & $0.73^{\pm 0.06}$ & $0.74^{\pm 0.03}$ & $\textbf{0.75}^{\pm 0.02}$ \\ 
		\bottomrule
	\end{tabular}
	
	\caption{Results L1C}
	\label{tab:resultsl1c}
\end{subtable}

\begin{subtable}{\textwidth}
	\begin{tabular}{rllllllll}
		\toprule
		  & shallow & \multicolumn{4}{c}{convolution} &  \multicolumn{2}{c}{recurrence} & \multicolumn{1}{c}{attention}
		  \\
		  \cmidrule(lr){2-2}\cmidrule(lr){3-6}\cmidrule(lr){7-8}\cmidrule(lr){9-9}
		  
		FRH04	& {RF} & {TempCNN} & {MS-ResNet} & {InceptionTime} & {OmniscCNN} & {LSTM} & {StarRNN} & {Transformer}  \\

overall accuracy & $0.78$ & $0.79$ & $0.77$ & $0.77$ & $0.73$ & $\textbf{0.80}$ & $0.79$ & $\textbf{0.80}$ \\ 
average accuracy & $0.54$ & $0.55$ & $0.54$ & $0.53$ & $0.52$ & $0.57$ & $0.56$ & $\textbf{0.58}$ \\ 
weighted f-score & $0.77$ & $0.79$ & $0.77$ & $0.77$ & $0.72$ & $\textbf{0.80}$ & $0.79$ & $\textbf{0.80}$ \\ 
kappa-metric & $0.71$ & $0.73$ & $0.70$ & $0.70$ & $0.65$ & $0.74$ & $0.73$ & $\textbf{0.75}$ \\ 
\midrule
FRH01, 02, 04
\\
overall accuracy & $0.78^{\pm 0.02}$ & $0.80^{\pm 0.01}$ & $0.77^{\pm 0.02}$ & $0.73^{\pm 0.04}$ & $0.77^{\pm 0.05}$ & $0.80^{\pm 0.02}$ & $0.80^{\pm 0.01}$ & $\textbf{0.81}^{\pm 0.01}$ \\ 
average accuracy & $0.54^{\pm 0.01}$ & $0.57^{\pm 0.01}$ & $0.57^{\pm 0.03}$ & $0.52^{\pm 0.01}$ & $0.55^{\pm 0.03}$ & $0.57^{\pm 0.01}$ & $0.56^{\pm 0.00}$ & $\textbf{0.59}^{\pm 0.01}$ \\ 
weighted f-score & $0.77^{\pm 0.02}$ & $0.80^{\pm 0.01}$ & $0.76^{\pm 0.01}$ & $0.69^{\pm 0.08}$ & $0.75^{\pm 0.06}$ & $0.80^{\pm 0.03}$ & $0.80^{\pm 0.01}$ & $\textbf{0.81}^{\pm 0.01}$ \\ 
kappa-metric & $0.71^{\pm 0.03}$ & $0.74^{\pm 0.01}$ & $0.71^{\pm 0.01}$ & $0.66^{\pm 0.05}$ & $0.70^{\pm 0.07}$ & $0.75^{\pm 0.03}$ & $0.74^{\pm 0.02}$ & $\textbf{0.76}^{\pm 0.02}$ \\ 

		\bottomrule
	\end{tabular}
	
	\caption{Results L2A}
	\label{tab:resultsl2a}
\end{subtable}
\caption{Accuracy metrics of all models benchmarked on the Breizhcrops dataset, considering L1C (a) and L2A (b) data. For each table, the top part displays the performance obtained when testing on the FRH04 region while training on the three remaining areas, whereas the bottom part displays average performance (plus one standard deviation) when models were tested on FRH01, FRH02 and FRH04 regions. Bold values show the highest performance.}
\label{tab:results}
\end{table*}


For the model selection, we trained models on FRH01 and FRH02 regions and chose hyperparameter values that gave the lowest validation loss calculated on the FRH03 region. All the model section procedure is conducted on top-of-atmosphere time-series data (L1C).
More precisely, we followed a two-step process where we first tuned the values of model-specific and optimization-specific hyperparameters for five epochs and then determined the number of optimal training epochs in a second step.

For the optimization-specific parameters, we sampled learning rate $\nu$ and weight decay $\lambda$ from the log-uniform distributions $\sim \mathcal{U}_\text{log}([10^{-2},10^{-4}])$ and $\sim \mathcal{U}_\text{log}([10^{-2}, 10^{-8}])$ for all models.
The model-specific hyperparameters varied per model.
For convolution-based models, we tested {TempCNN} models with convolution filter sizes $K \in \{3,5,7\}$ and a number of hidden representations $H \in \{2^5,2^6,2^7\}$. We also evaluated different dropout rates $d \sim \mathcal{U}([0,0.8])$. The only model-parameter to select for {MS-ResNet} is the number of hidden representations that we searched over $H \in \{2^5,2^6,2^7,2^8, 2^9\}$. Regarding the {InceptionTime} model, we evaluated the stacking of  $L \in \{1,2,3,4\}$ Inception modules with the following number of hidden representations $H \in \{2^5,2^6,2^7\}$. Finally, we did not tune any model-specific parameter for the OmniscaleCNN model since the authors do not recommend any tuning of the network.
For both {LSTM} and {StarRNN} recurrent-based networks, we evaluated models with several  cascaded layers $L \in \{1,2,3,4\}$, hidden vector dimensionalities of $H \in \{2^5,2^6,2^7\}$, and a dropout rate $d \sim \mathcal{U}([0,0.8])$. For the {LSTM} approach, we also tested mono-directional and bidirectional models.
Finally, we search over $H \in \{2^5,2^6,2^7,2^8, 2^9\}$ hidden representation with $N_\text{head} \in \{1,2,\dots, 8\}$ self-attention heads with a number of stacked layers $L$ ranging from 1 to 8 for {Transformer}. As per the other models, the dropout rate is drawn from a uniform distribution $d \sim \mathcal{U}([0,0.8])$.
This hyperparameter search was done via a random search for 12 hours on a DGX-1 server with a P-100 GPU for each model where three runs were trained in parallel per GPU with a batch-size of 256.

This first step of hyperparameter tuning resulted in a total of 25 OmniscaleCNN, 39 Transformer, 70 MS-ResNet,  78 TempCNN, 80 LSTM, 84 StarRNN and 86 InceptionTime trained models. For completeness, we detail below the selected configuration (\emph{i.e.}, the one that obtained the minimum validation loss) for each approach. {TempCNN} uses a kernel size $K$ of 7 with 128 hidden units $H$. The dropout rate is set to 18\%, the learning rate $\nu$ to $2.38 \cdot 10^{-4}$ and the weight decay $\lambda$ to $5.18 \cdot 10^{-5}$.
A {MSResNet} model composed of 32 hidden units $H$ with $\nu = 6.27 \cdot 10^{-7}$ and $\lambda=4.75 \cdot 10^{-6}$ was selected. InceptionTime stacks three Inception modules where each convolutional filter has a hidden vector dimensionality of $H=128$. The learning rate $\nu$ equals to $8.96 \cdot 10^{-3}$ and the weight-decay $\lambda$ to $2.22 \cdot 10^{-6}$. For {OmniscaleCNN}, we evaluated an optimal learning rate of $\nu = 1.06 \cdot 10^{-3}$ and a weight decay of $\lambda = 2.25 \cdot 10^{-7}$.
A bidirectional LSTM achieves the lowest validation loss by stacking 4 layers with 128 hidden units $H$. The dropout rate equals to 57\%, learning rate $\nu$ to $9.88 \cdot 10^{-3}$ and weight decay $\lambda$ to $5.26 \cdot 10^{-7}$.
A {StarRNN} model with 3 layers and an hidden vector dimensionality of $H=128$ with a learning rate $8.96 \cdot 10^{-3}$ and weight decay $\lambda = 2.22 \cdot 10^{-6}$ was optimal.
The {Transformer} achieved best validation performance with three layers and a single self-attention head, a vector dimensionality of $H=64$, 40\% dropout, $\nu = 1.31 \cdot 10^{-3}$ and $\lambda=5.52 \cdot 10^{-8}$. 

Given the model and optimization specific hyperparameters from above, we then search for the optimal number of epochs to train on.
Using the same experiment configuration, we retrain one model for each deep learning approach during 30 epochs on a GeForce RTX 2070. We monitor the validation loss and record the epoch number with the lowest value.
\cref{tab:runtime} displays the selected number of epochs for each approach.

\subsection{Results}
\label{subsec:result}

For the final model evaluation, we re-trained the models from scratch with their respective hyperparameter configuration.
We present the results in two formats in \cref{tab:results}.
In the top rows, we show the model performances on the FRH04 region that was completed in the previous hyperparameter tuning procedure.
For the latter rows, we followed a cross-evaluation scheme where we trained on three regions and evaluated the accuracy of the remaining one.
From the four possible folds, we discard the evaluation on FRH03 since the hyperparameters have been determined based on the performance of this region. 
We present mean and standard deviations from FRH01, FRH02, and FRH04 for the top-of-atmosphere and bottom-of-atmosphere data.

The evaluation is performed on a DGX-1 machine with one model per P-100 GPU which gave us the possibility to compare the runtime in iterations per second of one batch of 1024 data samples of each model, as summarized in \cref{tab:runtime}.
The runtime is similar between all models although the number of trainable parameters differs. In practice, we were able to train each model within less than 2 hours.

The attention-based transformer model \cite{Vaswani:transformer,russwurm2019self} slightly outperformed the recurrent models, \ie LSTM~\cite{russwurm2018multi} and StarRNN~\cite{turkoglu2019gating}, which obtained overall higher performance than Random Forests and convolution-based models with the exception of the TempCNN model \cite{pelletier2019temporal}.
This is consistent for both bottom-of-atmosphere (L1C) and top-of-atmosphere (L2A) data.

We hypothesize that the slightly worse performance of the convo\-lution-based approaches might be due to (1) the use of a different 
temporal sampling for each sample, (2) the inability of convolution-based models to deal with cloudy acquisitions.
Both experimental setting choices (applying a temporal subsampling and using cloudy information) could lead to a non-optimal learning process of discriminative convolution filters. Moreover, Temp\-CNN that obtains higher overall performance than MS-ResNet, Inception Time, and OmniscaleCNN is the only convolution-based model that has been specifically developed for land cover mapping. We leave a detailed evaluation of this systematic difference between model architectures to future research since it is beyond the scope of this paper.
We observed surprisingly little difference in model performances between top-of-atmosphere L1C data and bottom-of-atmosphere L2A data. 
However, we would like to emphasize that the hyperparameters have been determined on L1C data evaluated on the FRH03 region which may bias the results towards better L1C accuracies. We also leave further evaluations on the effectiveness of atmospheric correction to future research.

\section{Challenges}
\label{sec:challenges}

In the following, we outline a series of challenges associated with the dataset that pose demanding questions to the time series community and likely need to be addressed to improve the accuracy of methods trained on this dataset.

\textbf{Imbalanced class labels}. 
Agricultural areas are commonly dominated by few common crops, such as \textsl{corn}, \textsl{meadow}, or \textsl{wheat}, which are cultivated extensively.
Nevertheless, other types of vegetation are still of interest for the local authorities and should be classified at a reasonable accuracy.
This introduces a strong imbalance in the class frequencies, as shown in \cref{fig:classfrequencies}.
Please note the logarithmic scale. 

\textbf{Classification of noisy time series}.
Clouds cover the Earth's surface at irregular intervals and are inherent to all optical imagery. 
Their large reflectance introduces positive outliers to the data at single intervals which can be seen in the reflectance data over the time scale (see \cref{fig:example}).
Existing approaches classify, mask, and interpolate values from cloudy observations in a pre-processing step. 

\textbf{Regional variations in the class distributions}.
Regional variances in soil quality, elevation, temperature, and precipitation lead to a spatial correlation in the frequency of dominated agricultural crops.
This effect increases at larger scales where these environmental conditions change significantly.
Still, due to the nature of agricultural production focused on a few dominant crop types, a class imbalance can be observed in the data. 
Regional differences in environmental conditions further vary the label distribution for the respective partitions, as can be seen in the histogram of classes per region in \cref{fig:classfrequencies}.


\textbf{Variable sequence length}. 
Earth observation satellites scan the surface in stripes of \SI{290}{\kilo\meter} width (termed \textit{swath}).
To ensure a constant coverage, the acquisition is planned with a certain degree overlap towards the border of these stripes.
Due to this configuration, the sequence lengths $T$ of acquired images per field parcels vary between regions.

\textbf{Spatial autocorrelation}. 
Spatially close objects are more similar than distant ones \cite{tobler1970computer}. 
This autocorrelation can introduce a dependence between training and validation datasets that may disguise overfitting and impede generalization. 
To counteract this, several researchers \cite{russwurm2017temporal,jean2018tile2vec} have adopted a training/validation/evaluation partitioning that groups spatially distant parcels.
Hence, we organized the data in their respective NUTS-3 regions to encourage training on these spatially separate regions.

\section{Reproducibility}
\label{sec:code}

With \ourdataset, we aim at comparing the state of the art in crop type mapping. We thus release a curated code repository of labeled data, models, experiments, and evaluations in \gitrepo.
This Python package can be easily installed with
\mintinline{bash}{pip install breizhcrops}.

We provide here a minimal working example 
\begin{minted}{python}
    from breizhcrops import BreizhCrops, models
    x, y, field_id = BreizhCrops("frh04")[0]
    model = models.pretrained("Transformer")
    y_pred = model(x.unsqueeze(0))
\end{minted}
that downloads the FRH04 dataset and retrieves the first sample. Then, it loads a pretrained 
model (on FRH01, FRH02 and FRH03) and performs a prediction for this sample.

We believe that the accessibility to crop type data, classification models, and evaluation routines will accelerate developments in the scientific community and we welcome code contributions to include novel developments in the field of crop type mapping.

\section{Conclusion}
\label{sec:conlusion}

In this work, we presented a novel benchmark dataset, \ourdataset, for crop type mapping with top and bottom-of-atmosphere reflectance Sentinel-2 time series.
We evaluated seven recently developed state-of-the-art deep learning models on time series classification for crop type mapping along with a Random Forest classifier.
The attention-based Transformer model slightly outperformed the recurrent neural networks followed by the convolutional neural networks.
We release the dataset, model implementations and pre-trained model weights in an associated Python package that can be installed and run with few lines of code.
We hope that the accessibility of the dataset and deep learning models will encourage the community to benchmark novel crop type mapping methods on \ourdataset.
We encourage active code contributions to reflect the state of the art in crop type mapping with this dataset.
In future versions, we aim to include spatio-temporal models and label data from subsequent years.

\begin{spacing}{1}
	\normalsize
	\bibliography{references}
\end{spacing}

\end{document}